\documentclass{ecai}
\usepackage{graphicx}
\usepackage{latexsym}
\usepackage{amsmath}
\usepackage{algorithm}
\usepackage{algorithmic}
\usepackage{booktabs}
\usepackage{multirow}

\ecaisubmission   

\begin{document}

\begin{frontmatter}

\title{NeSyFOLD: Neurosymbolic Framework for Interpretable Image Classification}

\author[A]{\fnms{Parth}~\snm{Padalkar}\orcid{0000-0003-1015-0777}\thanks{Corresponding Author. Email: parth.padalkar@utdallas.edu.}}
\author[A]{\fnms{Huaduo}~\snm{Wang}\orcid{0000-0002-2118-5425}}
\author[A]{\fnms{Gopal}~\snm{Gupta}\orcid{0000-0001-9727-0362}} 

\address[A]{University of Texas at Dallas}

\begin{abstract}
Deep learning models such as CNNs have surpassed human performance in computer vision tasks such as image classification. However, despite their sophistication, these models lack interpretability which can lead to biased outcomes reflecting existing prejudices in the data. We aim to make predictions made by a CNN interpretable. Hence, we present a novel framework called NeSyFOLD to create a neurosymbolic (NeSy) model for image classification tasks. The model is a CNN with all layers following the last convolutional layer replaced by a stratified answer set program (ASP). A rule-based machine learning algorithm called FOLD-SE-M is used to derive the stratified answer set program from binarized filter activations of the last convolutional layer. The answer set program can be viewed as a rule-set, wherein the truth value of each predicate depends on the activation of the corresponding kernel in the CNN. The rule-set serves as a global explanation for the model and is interpretable. A justification for the predictions made by the NeSy model can be obtained using an ASP interpreter. 
We also use our NeSyFOLD framework with a CNN that is trained using a sparse kernel learning technique called Elite BackProp (EBP). This leads to a significant reduction in rule-set size without compromising accuracy or fidelity thus improving scalability of the NeSy model and interpretability of its rule-set. Evaluation is done on datasets with varied complexity and sizes. To make the rule-set more intuitive to understand, we propose a novel algorithm for labelling each kernel's corresponding predicate in the rule-set with the semantic concept(s) it learns. We evaluate the performance of our ``semantic labelling algorithm" 
to quantify the efficacy of the semantic labelling for both the NeSy model and the NeSy-EBP model.

\end{abstract}

\end{frontmatter}
\section{Introduction}
Interpretability in AI is an important issue that has resurfaced in recent years as deep learning models have become larger and are applied to an increasing number of tasks. Some applications such as autonomous vehicles \cite{kanagaraj2021}, disease diagnosis \cite{Sun2016ComputerAL}, and natural disaster prevention \cite{Ko2012disaster} are very sensitive areas where a wrong prediction could be the difference between life and death. The above tasks rely heavily on good image classification models such as Convolutional Neural Networks (CNNs). A CNN is a deep learning model used for a wide range of image classification and object detection tasks, first introduced by Y. Lecun et al. \cite{cnn}. Current CNNs are extremely powerful and capable of outperforming humans in image classification tasks. A CNN is inherently a blackbox model though attempts have been made to make it more interpretable \cite{zhang2017growing,zhang2018interpreting}.

While these deep models achieve remarkable accuracy, their interpretability becomes compromised. It remains unclear what the model has truly learnt and whether its predictions are rooted in human-understandable, semantically meaningful patterns or mere coincidental correlations within the dataset. Motivated by the need for interpretability, in this paper we introduce a framework called NeSyFOLD, which \textit{a)} gives a global explanation for the predictions made by a CNN in the form of an ordered rule-set and \textit{b)} generates an interpretable NeSy model that can be used to make predictions instead of the CNN. This resultant rule-set can be readily scrutinized by domain experts, enabling the discernment of potential biases towards specific classes or learning of non-intuitive class representations by the trained model.

\begin{figure}[t]
    \centering
    \includegraphics[width=\linewidth, height = 6cm]{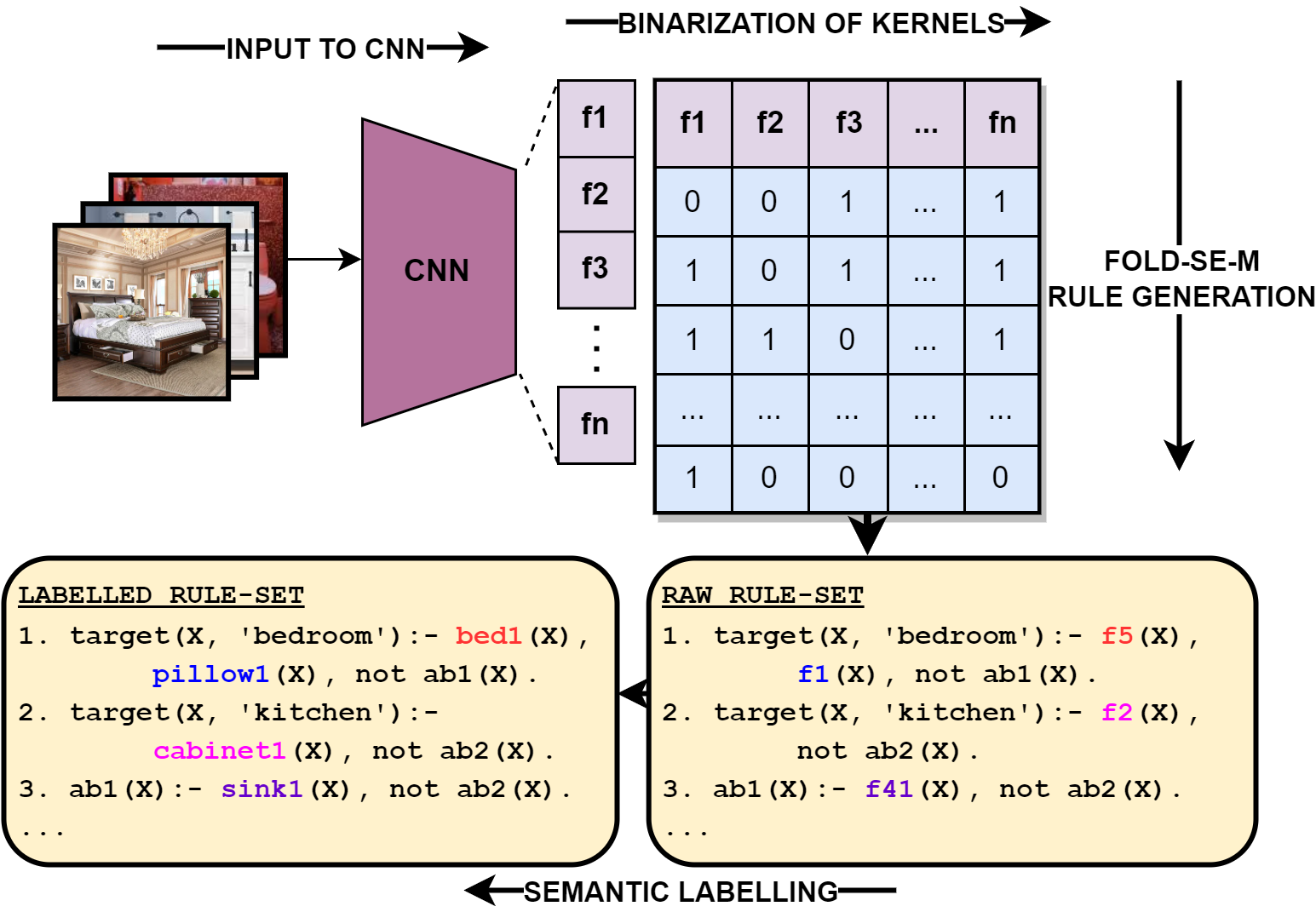}
    \caption{The NeSyFOLD Framework}
    \label{fig_1}
\end{figure}
The NeSyFOLD framework uses a Rule Based Machine Learning (RBML) algorithm called FOLD-SE-M \cite{foldsem} for generating a rule-set by using binarized outputs of the last layer kernels of a trained CNN. The rule-set is a (stratified) answer set program, i.e., Prolog extended with negation-as-failure. The most influential kernels, called \emph{significant} kernels, appear as predicates in the rule body.
The binarized kernel's output determines the truth value of its corresponding predicate in the rule-set. Aditionally, Zhou et al. \cite{objectdetectorfilters} showed that training a CNN on scene classification datasets results in last-layer kernels emerging as object detectors. Motivated by their discovery, we developed an algorithm for labelling the corresponding predicates of the \emph{significant} kernels with the semantic concept(s) that they represent. We call this procedure the \textit{semantic labelling} of the kernels. For example, the predicate {\tt 52(X)} corresponding to kernel {\tt 52} will be replaced by {\tt bathtub(X)} in the rule-set, if kernel {\tt 52} has learnt to look for ``bathtubs" in the image. Fig. \ref{fig_1} shows an outline of our NeSyFOLD framework. The semantic labelling can be done manually as demonstrated by Townsend et al. \cite{eric}, however, it entails substantial time and labor investments. We automate this process using our semantic labelling algorithm.

Next, we create a model that uses the generated rule-set in conjunction with the CNN for inference. We call this the \textit{NeSy} model. We use the in-built FOLD-SE-M rule interpreter \cite{foldsem} to execute the rules against the binarized vector of an image instance obtained from the last convolutional layer kernels of the CNN for predicting its class. To obtain the justification of a particular prediction we use the s(CASP) ASP solver \cite{arias_carro_salazar_marple_gupta_2018}. The justification serves as an explanation for the predictions made by the CNN. 

We compare our NeSyFOLD framework with the ERIC system \cite{eric} (SOTA) which also generates global explanations by extracting a rule-set in the form of a list of CNFs from the CNN through the use of a decision-tree like algorithm. 
 
We show through our experiments that our NeSy model 
outperforms ERIC's accuracy and fidelity while generating a significantly smaller-sized rule-set. The rule-set size can be further reduced by learning sparse kernels. We employ the Elite BackProp (EBP) \cite{EBP} algorithm to train the CNN so as to learn sparse kernels. We then obtain the NeSy-EBP model from this EBP-trained CNN and evaluate its performance as well.
Our novel contributions are as follows:

\begin{enumerate}
    \item We show that our method of rule extraction significantly reduces the size of the rule-set obtained, hence making the rule-set more intuitive and interpretable.
    \item We show that the scalability of our framework can be further improved by using sparse kernel learning techniques such as EBP.
    \item We present a novel algorithm for the semantic labelling of the predicates in the rule-set and do an extensive evaluation to show its effectiveness.
\end{enumerate}



\section{Background}
\medskip\noindent\textbf{FOLD-SE-M:}
The FOLD-SE-M algorithm \cite{foldsem} that we employ in our framework, learns a rule-set from data as a \textit{default theory}. 
Default logic is a non-monotonic logic used to formalize commonsense reasoning. A default $D$ is expressed as:
  
\begin{equation}\label{eq_1}
    D = A: \textbf{M} B \over\Gamma    
\end{equation}

\noindent Equation \ref{eq_1} states that the conclusion $\Gamma$ can be inferred if pre-requisite $A$ holds and $B$ is justified. $\textbf{M} B$ stands for ``it is consistent to believe $B$".
Normal logic programs can encode a default theory quite elegantly \cite{gelfondkahl}. A default of the form: 
$$\alpha_1 \land \alpha_2\land\dots\land\alpha_n: \textbf{M} \lnot \beta_1, \textbf{M} \lnot\beta_2\dots\textbf{M}\lnot\beta_m\over \gamma$$
\noindent can be formalized as the
normal logic programming rule:
$$\gamma ~\texttt{:-}~ \alpha_1, \alpha_2, \dots, \alpha_n, \texttt{not}~ \beta_1, \texttt{not}~ \beta_2, \dots, \texttt{not}~ \beta_m.$$
\noindent where $\alpha$'s and $\beta$'s are positive predicates and \texttt{not} represents negation-as-failure. We call such rules \emph{default rules}. 
Thus, the default 

$$bird(X): M \lnot penguin(X)\over flies(X)$$

\noindent will be represented as the following default rule in normal logic programming:

{\tt flies(X) :- bird(X), not penguin(X).}

\noindent We call {\tt bird(X)}, the condition that allows us to jump to the default conclusion that {\tt X} flies, the {\it default part} of the rule, and {\tt not penguin(X)} the \textit{exception part} of the rule. 

FOLD-SE-M \cite{foldsem} is a Rule Based Machine Learning (RBML) algorithm. It generates a rule-set from tabular data, comprising rules in the form described above. The complete rule-set can be viewed as a stratified answer set program. It uses special {\tt abx} predicates to represent the exception part of a rule where {\tt x} is unique numerical identifier.   
FOLD-SE-M incrementally generates literals for \textit{default rules} that cover positive examples while avoiding covering negative examples. It then swaps the positive and negative examples and calls itself recursively to learn exceptions to the default when there are still negative examples falsely covered.

There are $2$ tunable hyperparameters, $ratio$, and $tail$. 
The $ratio$ controls the upper bound on the number of false positives to the number of true positives implied by the default part of a rule. The $tail$ controls the limit of the minimum number of training examples a rule can cover.
FOLD-SE-M generates a much smaller number of rules than a decision-tree classifier and gives higher accuracy in general.

\medskip\noindent\textbf{Elite BackProp:}
Elite BackProp \cite{EBP} is a training algorithm that associates each class with very few \textit{Elite} kernels that activate strongly. This is achieved by introducing a loss function that penalises the kernels that have a lower probability of activation for any class and reinforces the kernels that have a higher probability of activation during training. This leads to fewer kernels learning representations for each class. The same number of (elite) kernels are reinforced for each class which is decided by the hyperparameter $K$. Another hyperparameter $\lambda$ is used as the regularization constant.

\section{Methodology}
We now proceed to explain the methodology behind the learning, inference, and semantic labelling pipeline used in our NeSyFOLD framework.

\subsection{Learning}
We start by training the CNN on the input images for the given classification dataset. Any optimization technique can be used for updating the weights. The learning pipeline is illustrated in Figure \ref{fig_1}.

\medskip\noindent\textbf{Quantization:}
Quantization is the process of binarization of the kernel outputs.
Once the CNN has been fully trained to convergence, we pass the full train set consisting of $n$ images to the CNN. For every image $i$ in the train set we obtain a feature map $A_{i,k}$ for every kernel $k$ in the last convolutional layer. The feature map $A_{i,k}$ is a $2D$ matrix of dimension determined by the CNN architecture. For each image $i$ there are $K$ feature maps generated where $K$ is the total number of kernels in the last convolutional layer of the CNN.
We map each of these feature maps to a single real value by taking their norms as demonstrated by eq. \eqref{eq_2}.
Next, to determine an appropriate threshold $\theta_k$ for each kernel $k$ to binarize its output, we use a weighted sum of the mean and the standard deviation of the norms $a_{i,k}$ for all $i$ for a given $k$ using eq. \eqref{eq_3} where $\alpha$ and $\gamma$ are hyperparameters. 
Binarizing a kernel refers to determining whether the kernel is active or inactive for a specific image. Thus a binarization table $B$ is created. Each row in the table represents an image and each column is the binarized kernel feature map value represented by either a $0$ (inactive) or $1$ (active). This is done using eq. \eqref{eq_4} where $Q(A_{i,k}, \theta_k)$ is the quantization function that outputs a value of $1$ if, for a feature map $A_{i,k}$ its norm $a_{i,k}$ is greater than the kernel $k's$ threshold $\theta_k$ and $0$ otherwise.

\begin{align}
    a_{i,k} =& ||A_{i,k}||_2 \label{eq_2}\\
    \theta_{k} = & \alpha \cdot \overline{a_{k}} + \gamma\sqrt{\frac{1}{n}\sum(a_{i,k} - \overline{a_{k}})^2} \label{eq_3}\\
    Q(A_{i,k},\theta_k) = & 
    \begin{cases}
    1, & \text{if } a_{i,k} > \theta_k\\
    0,   & \text{otherwise}\\
    \end{cases}\label{eq_4}
\end{align}

\medskip\noindent\textbf{Rule-set Generation:} The binarization table $B$ is given as an input to the FOLD-SE-M algorithm to obtain a rule-set in the form of a stratified answer set program. The raw rule-set has predicates with names in the form of their corresponding kernel's index.
An example rule could be:\\
{\tt target(X,`2') :- not 3(X), 54(X), 

~~~~~~~~~~~~~~~~not ab1(X).}\\
This rule can be interpreted as ``Image X belongs to class {\tt`2'} if kernel {\tt 3} is not activated
and kernel {\tt 54} is activated and the abnormal condition (exception) {\tt ab1} does not apply". There will be another rule with the head as {\tt ab1(X)} in the rule-set. The binarized output of a kernel would determine the truth value of its predicate in the rule-set.

\medskip\noindent\textbf{Semantic labeling :} Every kernel in the last layer of the CNN learns to identify certain concepts. Since we capture the outputs of the kernels as truth values of predicates in the rule-set, we can label the predicates as the semantic concept(s) that the corresponding kernel has learnt. 
Thus, the same example rule from above may now look like:\\
~~~~~~{\tt target(X,`bathroom') :- not bed(X),  

  ~~~~bathtub(X), not ab1(X).}
  
Currently, the process of attributing semantic labels to predicates involves identifying \textit{top-m} images with the highest feature-map norms generated by the kernel. Next, the assignment of semantic labels is executed through manual observation of the resulting $m$ images by discerning the concepts that emerge most prominently. This methodology is exemplified in the work of Townsend et al. \cite{eric}. We introduce a novel semantic labelling algorithm to automate the semantic labelling of the predicates. The details of the algorithm are discussed later.

\subsection{Inference}
The inference pipeline of NeSyFOLD is relatively straightforward. We feed the test set images to the CNN to obtain the kernel feature maps for each kernel in the last convolutional layer. Then, using eq. \eqref{eq_4}, we get the binarizations for each kernel output and generate the binarization table $B_{test}$. Note here that the threshold $\theta_k$ for each kernel $k$ is the same that was calculated in the learning process. Next, for each binarized vector $b$ in $B_{test}$, we run the FOLD-SE-M toolkit's built-in rule interpreter on the labeled/unlabelled rule-set to make predictions. The truth value of the predicates in the rule-set is determined by the corresponding binarized kernel values in $b$. The binarized kernel values in $b$ can be listed as facts and the ASP rule-set can be queried with s(CASP) interpreter to obtain the justification as well as the target class. The aforementioned procedure can be conceptualized as replacing all the layers following the last convolutional layer with the rule-set and then making the predictions. This integrated model is referred to as the NeSy model. 

\subsection{Semantic Labelling of Significant Kernels}
Recall that a \textit{significant} kernel is a kernel whose corresponding predicate appears in the rule-set generated by FOLD-SE-M.
The rule-set initially has kernel ids as predicate names.  Also, the FOLD-SE-M algorithm finds only the most influential kernels and uses them as predicates. We present a novel algorithm for automatically labelling the corresponding predicates of the \emph{significant} kernels with the semantic concept(s) that the kernels represent as shown in Alg. \ref{alg_1}.

\begin{figure}[t]
    \centering
    \includegraphics[width=\linewidth, height = 5cm]{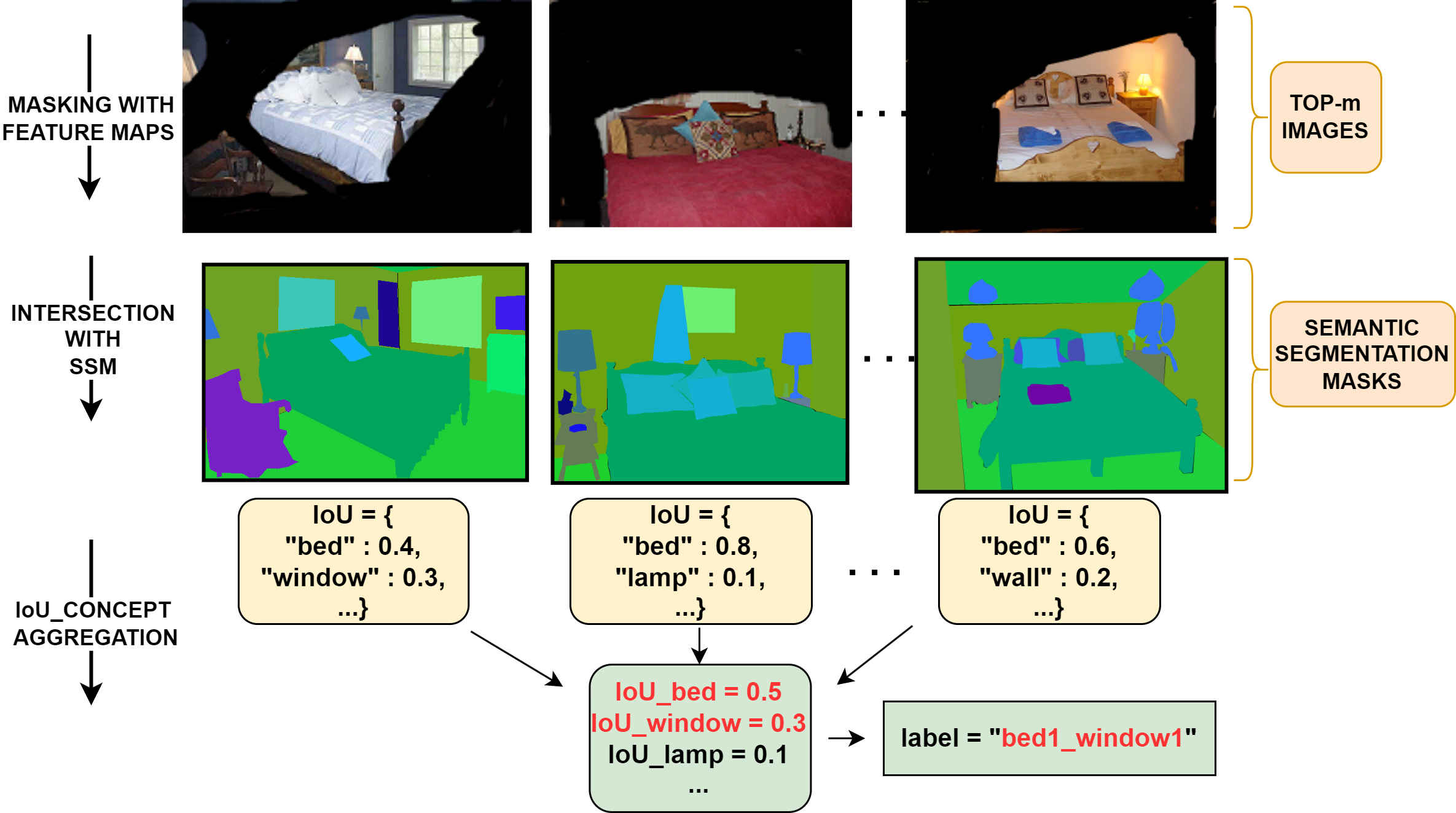}
    \caption{Semantic labelling of a predicate ($margin = 0.2$).}
    \label{fig_2}
\end{figure}

Xie et al. \cite{conceptsinCNN} showed that each kernel in the CNN may learn to represent multiple concepts in the images.  As a result, we assign semantic labels to each predicate, denoting the names of the semantic concepts learnt by the corresponding kernel. 
To regulate the extent of approximation, i.e. to dictate the number of concept names to be included in the predicate label, we introduce a hyperparameter 
\textit{margin}.
This hyperparameter exercises control over the precision of the approximation achieved. Figure \ref{fig_2} illustrates the semantic labelling of a given predicate. We use the \textit{ADE20k} dataset \cite{ade20k} in our experiments. It provides manually annotated semantic segmentation masks for a few images of all the classes of the Places \cite{zhou2017places} and SUN \cite{sun} datasets. This essentially means that for every image $i$ in the dataset $I$, there is an image $i_M$ where every pixel is annotated with the label of the object (concept) that it belongs to (Fig. \ref{fig_2} middle). In Alg. \ref{alg_1} we denote these by $I_M$.

The CNN $M$ that is trained on the train set is used to obtain the norms $a_{i,k}$ of the feature maps $A_{i,k}$ extracted from the last convolution layer for each kernel $k$ same as in the learning process.
Now for each kernel $k$ the \textit{top-m} images $I_m$, according the norms $a_{i, k}$ are selected. These images are masked with their resized feature maps $A_{i, k}$ to obtain $\hat{I}_m$ (Fig. \ref{fig_2} top).
Next, for each masked image $\hat{i}_m \in \hat{I}_m$ we calculate the $IoU_c$ score (eq. \eqref{eq_5}) for each concept $c$ that appears in its corresponding semantic segmentation mask $i_{Mask}$ and sum the scores over all images $I_m$ for each concept $c$.
We then normalize the scores over all $c$ and sort the concepts in descending order w.r.t their respective $IoU_c$ scores.
Finally, the predicate associated with significant kernel $k$ is labelled as a concatenation of all the concepts that are in a specific margin from the concept with the highest $IoU_c$ score. This is determined by the \textit{margin} hyperparameter. For example, if the concept $IoU_c$ scores for kernel $12$ are \{$cabinets$ : $0.5$, $door$ : $0.4$, $drawer$ : $0.1$\} then with a \textit{margin} of $0.1$ the label for the kernel will be ``$cabinets1\_door1$". Also, if some kernel previously has already been labelled $cabinets1$, then the numerical identifier for the label for kernel $12$ will be $2$ for $cabinets$ in the label i.e. ``$cabinets2\_door1$". 
 \begin{align}
     IoU_c(i_{Mask}, \hat{i}) =& \frac{\text{no. of pixels in } c \cap \hat{i}}{\text{no. of pixels in } \hat{i}} \label{eq_5}
 \end{align}
\begin{algorithm}[h]
\caption{Semantic labelling}
\label{alg_1}
\begin{algorithmic}[1]
    \REQUIRE The trained CNN model $M$, the images $I$ and their semantic segmentation masks $I_{Mask}$, unlabelled rule-set $R$, hyperparameter $margin$
    
    \STATE Obtain feature maps $A_{i,k}$, norms $a_{i,k}$ for all significant kernels $k$ and images $i$ in $I$ using $M$
    \FOR{Each significant kernel $k$ in $R$}
    \STATE $I_m$ $\gets$ $top-m$ images w.r.t $a_{i,k}$
    \STATE $\hat{I}_{m}$ $\gets$ Mask each $i_m$ in $I_m$ with its resized $A_{i_{m}, k}$ 
    \FORALL{$\hat{i}_{m}$ in $\hat{I}_{m}$}
    \STATE $IoU^{\hat{i}_m}_c$ $\gets$ $IoU_c(i_{Mask}, \hat{i}_{m})$ for each concept $c$ in $i_{Mask}$ 
    \ENDFOR
    \STATE $IoU_c$ $\gets$ $\overset{{\hat{i}_{m}}}{\sum} IoU^{\hat{i}_m}_c$ for all $c$
    \STATE Reverse sort normalized $IoU$ scores over all $c$
    \STATE Label the kernel $k$ as all concepts $c$ that have $IoU$ score within a $margin$ from the top concept. 
    \ENDFOR
\end{algorithmic}
\end{algorithm}

\section{Experiments}
We conducted experiments to address the following questions:

\medskip\noindent\textbf{Q1:} How well does NeSyFOLD scale w.r.t number of classes and number of images in the train set?

\medskip\noindent\textbf{Q2:} What is the effect of using a CNN trained with Elite BackProp in our NeSyFOLD framework?

\medskip\noindent\textbf{Q3:} How well do the semantic labels associated with the predicates adhere to the concepts during inference?

\medskip\noindent\textbf{Q4:} How many concepts are the predicate labels comprised of after semantic labelling in general?

\medskip\noindent\textbf{[Q1/Q2] Scalability:} We define scalability as a function of accuracy, fidelity and size of 
rule-set. As the number of classes in the dataset increases, the accuracy and fidelity of the framework should be high w.r.t the trained CNN and the rule-set generated should be as small as possible to increase interpretability \cite{rulesetinterpretability}. To determine the scalability of our NeSyFOLD framework we evaluate the fidelity, accuracy, no. of unique predicates/atoms in the rule-set and the overall rule-set size over various subsets of the \emph{Places} \cite{zhou2017places} dataset which has images of various scenes and the \emph{German Traffic Sign Recognition Benchmark} (GTSRB) \cite{gtsrb} dataset which consists of images of various traffic signposts. We selected subsets of $2$, $3$, $5$ and $10$ classes from the \emph{Places} dataset. We started with the bathroom and bedroom classes (P2). Subsequently, we incorporated the kitchen class (P3.1), followed by the dining room and living room (P5), and finally, home office, office, waiting room, conference room, and hotel room (P10). We also selected $2$ additional subsets of $3$ classes each i.e. {desert road, forest road, street} (P3.2) and {desert road, driveway, highway} (P3.3). We show the evaluation results of our semantic labelling algorithm on P3.1, P3.2 and P3.3 later. Each class has $5k$ images of which we made a $4k/1k$ train-test split for each class and we used the given validation set as it is. The \textit{GTSRB} (GT43) dataset has $43$ classes of signposts. We used the given test set of $12.6k$ images as it is and did an $80:20$ train-validation split which gave roughly $21k$ images for the train set and $5k$ for the validation set.

We use rule-set size as a metric of interpretability. Lage et al. \cite{rulesetinterpretability} showed through human evaluations that as the size of the rule-set increases the difficulty in interpreting the rule-set also increases. 
Size is calculated as the total number of antecedents for ERIC and the total number of predicates in the bodies of the rules generated by NeSyFOLD.

\noindent\textbf{Setup}: We employed a VGG16 CNN pretrained on $Imagenet$ \cite{deng2009imagenet}, training over $100$ epochs with batch size $32$. The Adam \cite{adamoptim} optimizer was used, accompanied by class weights to address data imbalance. Regularization of $L2$ at $0.005$ spanning all layers, and a learning rate of $5 \times 10^{-7}$ was adopted. A decay factor of $0.5$ with a $10$-epoch patience was implemented. Images were resized to $224 \times 224$, and hyperparameters $\alpha$ and $\gamma$ (eq. \eqref{eq_3}) for calculating threshold for binarization of kernels, were set at $0.6$ and $0.7$ respectively.

We re-trained each of the trained CNN models again for each dataset, for $50$ epochs using EBP. We used $K = 20$ for ``P10" and ``GT43" because of their larger size and $K=5$ for all the other datasets. We used $\lambda = 0.005$ for all datasets. We chose EBP for learning sparse kernels because Kasioumis et. al \cite{EBP} show that EBP learns better representations than other sparse kernel learning techniques. We then used our NeSyFOLD framework both with the vanilla-trained CNN and the EBP-trained CNN to generate the respective NeSy models (NF and NF-E) as described previously. 
For P10, per run, we selected the highest $10\%$ softmax-scored images in each class. This subset was then used exclusively for rule-set generation. Due to the CNN's lower accuracy on PLACES10, kernel representations suffered, contributing noise to binarized kernels. This technique counteracted this noise, elevating NeSy model's accuracy and fidelity.
The performance of these NeSy models are reported in Table \ref{tb_1}. The results are reported after 5 runs on each dataset. For comparison, the trained CNN's accuracy is $\{0.97, 0.94, 0.96, 0.89, 0.85, 0.70, 0.98\}$ for the datasets listed in Table \ref{tb_1} in the same order.  

Our experimental setup closely resembled Townsend et al.'s \cite{townsend2022on}, \cite{eric} facilitating a meaningful comparison. The performance metrics of ERIC reported in Table \ref{tb_1} are also taken from these papers.
ERIC's fidelity for P3.2 and P3.3 remains unreported in the mentioned papers and thus is left blank in the table. 
Given space constraints, comprehensive hyperparameter values are detailed in the supplementary material.

\begin{table*}[ht]
\centering
\setlength{\tabcolsep}{4.0pt}
\begin{tabular}{@{}rlllll@{}}
\toprule
\multicolumn{1}{l}{Data} & Algo & Fid. & Acc.& Pred. & Size \\ \midrule
\multirow{3}{*}{P2}      & ERIC &$0.89 \pm 0.01$ & $0.89 \pm 0.01$ & $11 \pm 1$ & $52 \pm 8$\\
                            & NF   & $ \textbf{0.93} \pm \textbf{0.01} $ &$ \textbf{0.92} \pm \textbf{0.01} $ & $16 \pm 2$ & $28 \pm 5$\\
                            & NF-E & $ \textbf{0.93} \pm \textbf{0.01} $ &$ \textbf{0.92} \pm \textbf{0.01} $ & $ \textbf{8} \pm \textbf{2} $ & $ \textbf{12} \pm \textbf{5} $\\ \cmidrule(lr){1-6}
\multirow{3}{*}{P3.1}      & ERIC &$0.82 \pm 0.01$ & $0.81 \pm 0.01$ & $33 \pm 4$ & $118 \pm 13$\\
                            & NF   & $0.85 \pm 0.03$ &$0.84 \pm 0.03$ & $28 \pm 6$ & $49 \pm 9$ \\
                            & NF-E & $ \textbf{0.87} \pm \textbf{0.02} $ &$ \textbf{0.86} \pm \textbf{0.02} $ & $ \textbf{10} \pm \textbf{3} $ & $ \textbf{16} \pm \textbf{5} $\\ \cmidrule(lr){1-6}

\multirow{3}{*}{P3.2}      & ERIC &- & $0.90$ & $33$ & $127$\\
                            & NF   & $ \textbf{0.94} \pm \textbf{0.0} $ &$ \textbf{0.92} \pm \textbf{0.0} $ & $16 \pm 4$ & $26 \pm 7$\\
                            & NF-E & $ 0.92 \pm 0.01 $ &$ 0.91 \pm 0.01 $ & $ \textbf{6} \pm \textbf{1} $ & $ \textbf{7} \pm \textbf{1} $\\ \cmidrule(lr){1-6}
\multirow{3}{*}{P3.3}      & ERIC & - & $0.75$ & $44$ & $176$\\
                            & NF   & $\textbf{0.83} \pm \textbf{0.01}$ &$\textbf{0.79} \pm \textbf{0.01}$ & $32 \pm 5$ & $60 \pm 11$\\
                            & NF-E & $0.82 \pm 0.03$ &$0.78 \pm 0.03$ & $ \textbf{9} \pm \textbf{0} $  & $ \textbf{23} \pm \textbf{4} $\\ \cmidrule(lr){1-6}
\multirow{3}{*}{P5}      & ERIC &$0.65 \pm 0.01$ & $0.63 \pm 0.01$ & $57 \pm 4$ & $171 \pm 10$\\
                            & NF   & $0.67 \pm 0.03$ &$0.64 \pm 0.03$ & $56 \pm 3$ & $131 \pm 10$\\
                            & NF-E & $ \textbf{0.70} \pm \textbf{0.02} $ &$ \textbf{0.67} \pm \textbf{0.02} $ & $ \textbf{12} \pm \textbf{4} $ & $ \textbf{30} \pm \textbf{8} $\\ \cmidrule(lr){1-6}                      
\multirow{3}{*}{P10}      & ERIC &$0.39 \pm 0.01$ & $0.36 \pm 0.01$ & $85 \pm 6$ &  $208 \pm 16$ \\
                            & NF   & $0.46 \pm 0.02$ &$0.42 \pm 0.02$ & $84 \pm 6$ & $131 \pm 11$\\
                            & NF-E & $ \textbf{0.49} \pm \textbf{0.02} $ &$ \textbf{0.44} \pm \textbf{0.01} $ & $ \textbf{36} \pm \textbf{4} $ & $ \textbf{65} \pm \textbf{9} $\\ \cmidrule(lr){1-6}
\multirow{3}{*}{GT43}      & ERIC &$0.73 \pm 0.01$ & $0.73 \pm 0.01$ & $232 \pm 3$ & $626 \pm 28$\\
                            & NF   & $0.75 \pm 0.04$ &$0.75 \pm 0.04$ & $206 \pm 28$ & $418 \pm 79$ \\
                            & NF-E & $ \textbf{0.85} \pm \textbf{0.13} $ &$ \textbf{0.85} \pm \textbf{0.13} $ & $ \textbf{58} \pm \textbf{14} $ & $ \textbf{99} \pm \textbf{28} $\\ \cmidrule(lr){1-6}
\\\cmidrule(lr){1-6}                            
\multirow{3}{*}{\textbf{MS}}      & ERIC & $0.70 \pm 0.01$ & $0.72 \pm 0.01$ & $71 \pm 4$ & $211 \pm 15$\\
                            & NF   & $0.78 \pm 0.02$ &$0.75 \pm 0.02$ & $63 \pm 8$ & $120 \pm 19$ \\
                            & NF-E & $ \textbf{0.80} \pm \textbf{0.03} $ &$ \textbf{0.78} \pm \textbf{0.03} $ & $ \textbf{20} \pm \textbf{4} $ & $ \textbf{36} \pm \textbf{9} $ \\ \cmidrule(lr){1-6}

\end{tabular}
\caption{Comparison of ERIC vs NeSyFOLD (NF) vs NeSyFOLD-EBP (NF-E). \textbf{MS} shows the average value for each evaluation metric for each framework.}
\label{tb_1}
\end{table*}

Note that comparison is done only for the datasets used for evaluation of the ERIC system \cite{eric}, as the ERIC system code is proprietary.\\
\noindent\textbf{Result:} We observe that as the number of classes increases there is a drop in the fidelity (Fid.) and accuracy (Acc.) in general. This is because as more classes are introduced, more kernels in the CNN have to learn representations for those classes and hence there is more loss in the binarization of the kernels to obtain the rule-set.
Our NeSy model generated for both NeSyFOLD (NF) and NeSyFOLD-EBP (NF-E) outperforms ERIC on all of these datasets w.r.t accuracy and fidelity. All models perform poorly on the P10 dataset. We believe that this is because of the fewer distinct edges in the images in the P10 dataset which makes it harder for even the CNN kernels to learn good representations. GT43 has well-defined edges and consequently the accuracy and fidelity of all ERIC, NF and NF-E is high.

The size of the rule-set generated by NF is always smaller in size than ERIC. Moreover, NF-E generates an even smaller rule-set without compromising on accuracy or fidelity. Recall, the smaller the rule-set size the better the interpretability and lesser the manual semantic labelling effort. Since the number of Elite kernels is relatively small and defined for EBP, a smaller number of kernels learn tighter representations for each class. Consequently, there is less loss of information in binarization and the FOLD-SE-M algorithm finds a rule-set with a small number of predicates. Hence, using sparse kernel learning techniques such as EBP improves the scalability of NeSyFOLD. 

\medskip\noindent\textbf{[Q3/Q4] Semantic Labelling Efficacy:} 
To study the efficacy of the semantic labelling of the predicates in the rule-set using Alg. \ref{alg_1}, we selected the
P3.1, P3.2 and P3.3 datasets from the previous experiment. These datasets were chosen because NF and NF-E show decent accuracy and generate a comparatively smaller rule-set.\\
\noindent\textbf{Setup:}
The \textit{ADE20k} \cite{ade20k} dataset has manually annotated semantic segmentation masks of a few images of all the classes in P3.1, P3.1 and P3.3. We used these as input to the semantic labelling algorithm (Alg. \ref{alg_1}).\\  

\begin{figure}[]
    \centering
    \includegraphics[width=\linewidth, height = 8cm]{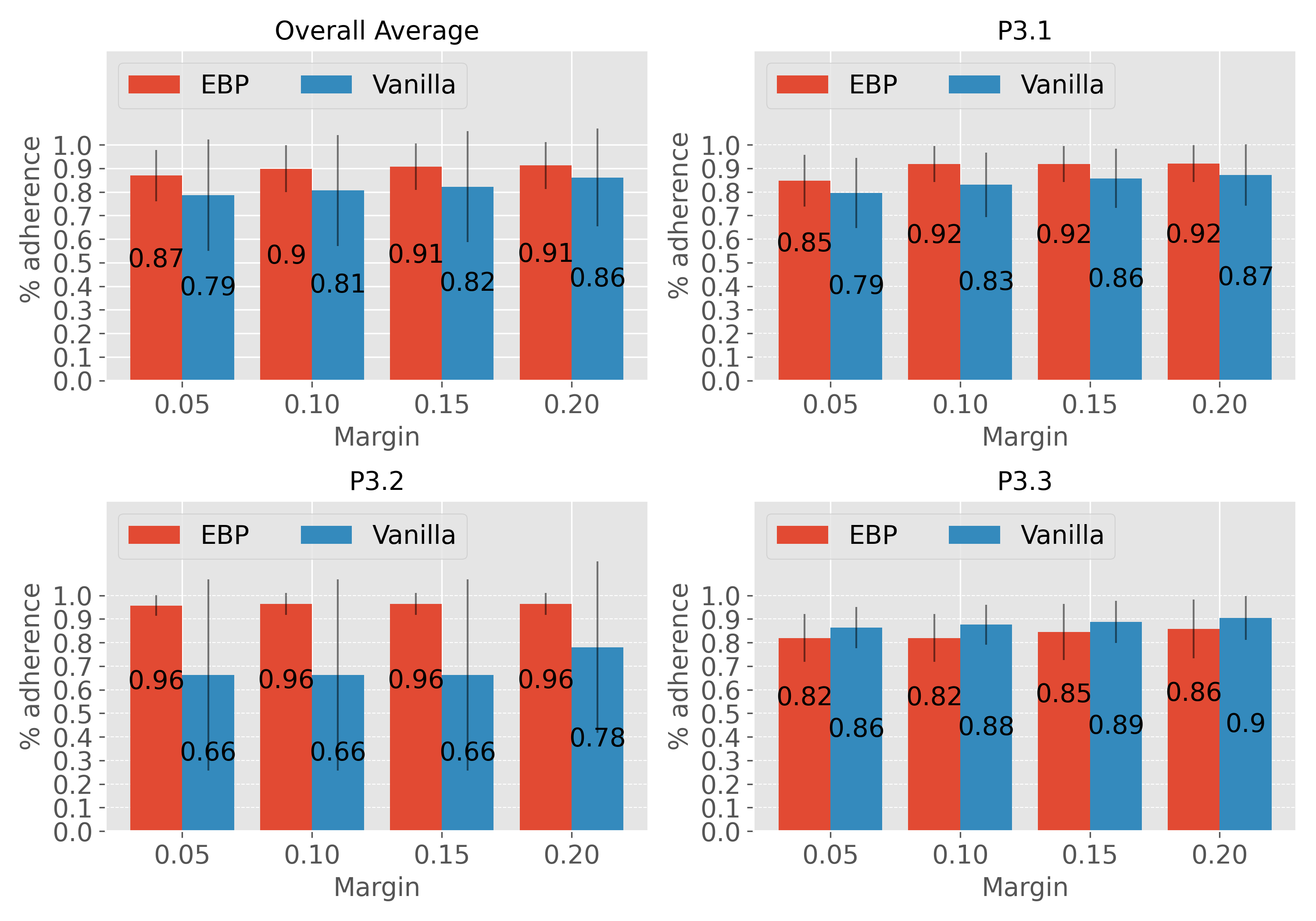}
    \caption{Percentage Adherence plots for kernels in NeSyFOLD-EBP (red) and NeSyFOLD (blue)}
    \label{fig_3}
\end{figure}

It is important that the kernel associated with a predicate adheres to the semantic label of the predicate. If it is not the case then the interpretability of the rule-set diminishes significantly.
To quantify \textit{adherence} of a kernel to the concepts that its corresponding predicate has been labelled with, we obtained the unlabelled rule-set for $1$ run of NeSyFOLD and NeSyFOLD-EBP on each of the datasets and labelled the rule-sets with $4$ different values $\{0.05, 0.1, 0.15, 0.2\}$ of \textit{margin} hyperparameter. We selected \textit{top-10} images according to their feature-map norms for each kernel. Subsequently, for test set images where the NeSy model predicts correctly, we identified the activated rule and extracted the ``True" predicates. We then manually scrutinize if any concepts from these predicates' labels are discernible within the image masked by the resized feature map originating from the relevant kernel. If even one concept in the label is adequately visible through the masked image we count that as positive. We say that the kernel \textit{adheres} to its given label for this image. Finally, we calculated the average percentage adherence of the kernels for each dataset, for all the $4$ \textit{margin} values and summarize the comparison between NeSyFOLD and NeSyFOLD-EBP in Fig \ref{fig_3}.

\begin{figure}[]
    \centering
    \includegraphics[width=\linewidth, height = 9cm]{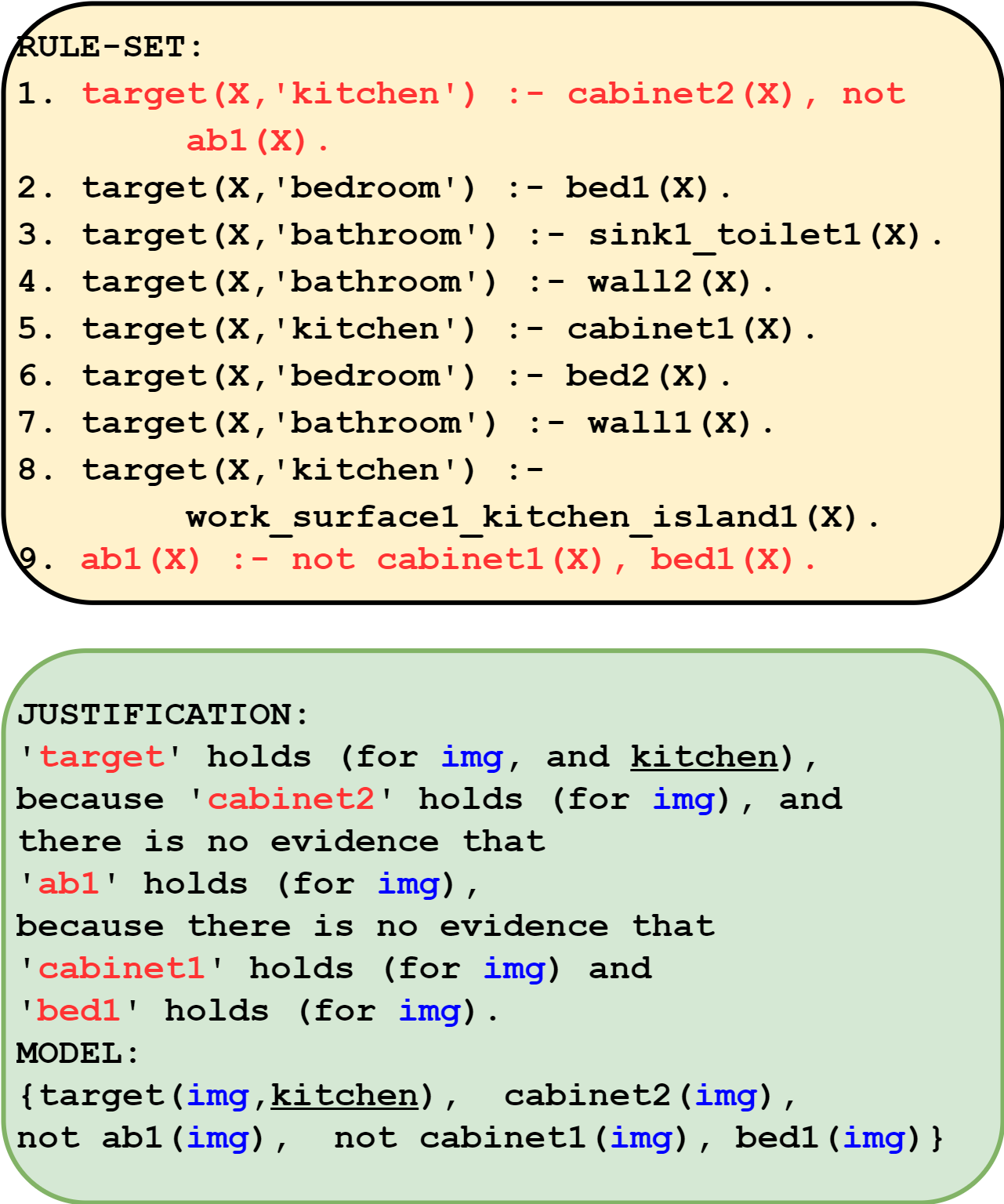}
    \caption{A rule-set generated by NeSyFOLD-EBP for P3 dataset (top). Justification generated by s(CASP) for an image ``img" for which the first rule was fired (bottom).}
    \label{fig_4}
\end{figure}
\begin{figure}[t]
    \centering
    \includegraphics[width=\linewidth, height =5cm]{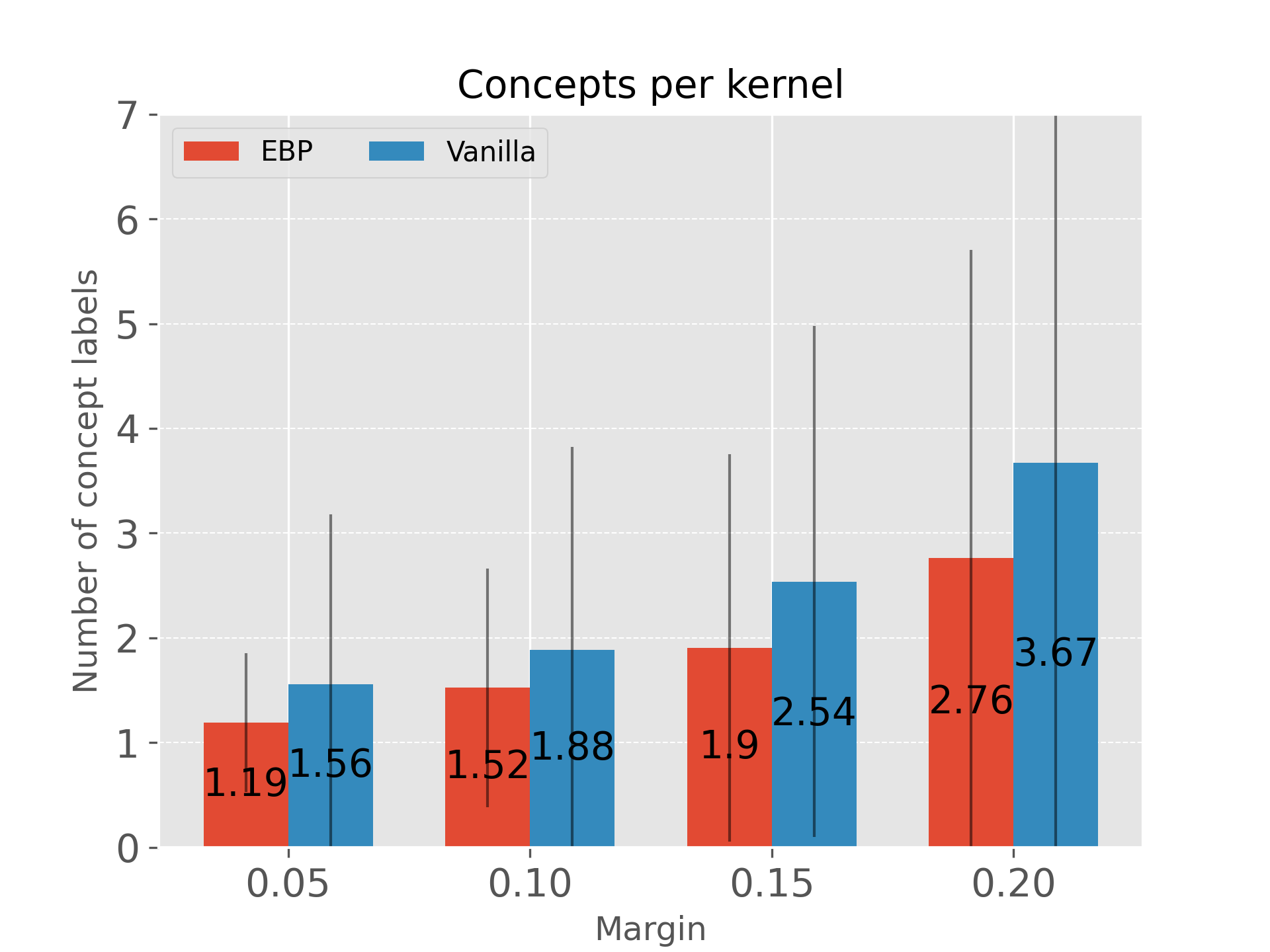}
    \caption{The average number of concept labels for kernels in NeSyFOLD-EBP (red) and NeSyFOLD (blue).}
    \label{fig_5}
\end{figure}

\noindent\textbf{Result:}
Recall that the \textit{margin} hyperparameter controls the level of approximation of the semantic labelling.
We observe that as the \textit{margin} value is increased, the average percentage adherence overall as well as for individual datasets, increases. This is intuitive because as the \textit{margin} value is increased the number of concepts appearing in the semantic label of each predicate also increases (Fig \ref{fig_5}). Hence, it is more likely that a  kernel adheres to a given image if its corresponding predicate has more concepts in its semantic label. 

Note that the kernels of the CNN trained with EBP show higher average percentage adherence overall. This is expected because the kernels learn tighter representations hence generating smaller receptive fields which focus on very few and specific concepts.

Figure \ref{fig_5} shows the average number of concepts appearing in the semantic label of the predicate as the \textit{margin} hyperparameter is varied. As expected, for NeSyFOLD-EBP the number of concepts is lower for every value of \textit{margin}. Also, the variation in the number of concepts is lesser.

Ideally, the number of concepts in the labels of the predicates should be low while the percentage adherence should be high for having interpretability with high confidence. NeSyFOLD-EBP shows promising results in this regard. In Figure \ref{fig_4} on the top we show a rule-set obtained using NeSyFOLD-EBP on the P3.1 dataset with a \textit{margin} of $0.05$. To determine the class of a given image ``img" we obtain its binarized vector, list it as facts in s(CASP) and run the query: {\tt ?- target(img, X).} The s(CASP) interpreter checks the rules from the top and returns a model that satisfies the ASP rule-set. On the bottom of Figure \ref{fig_4} we show a justification that can be obtained from the s(CASP) interpreter for the classification if the first rule (red) was the one that fired.

\section{Related Work}
There have been efforts since the 1990s to extract knowledge from a neural network. Andrews et al. \cite{ANDREWS1995373} have classified these efforts into three kinds: (i) \emph{pedagogical} methods wherein rules are constructed to explain the output in terms of the input, (ii) \emph{decompositional} methods which extract rule sets specific to different parts of the network, and (iii) \emph{eclectic} methods that are a combination of both. Both NeSyFold and ERIC are decompositional methods.

There is significant past work which focuses on visualizing the outputs of the layers of the CNN. These methods try to map the relationship between the input pixels and the output of the neurons. Zeiler et al. \cite{zeiler2014visualizing} and  Zhou et al. \cite{zhou2016learning} use the output activation while others \cite{selvaraju2017grad,https://doi.org/10.48550/arxiv.1412.6815,https://doi.org/10.48550/arxiv.1312.6034} use gradients to find the mapping. Unlike NeSyFold, these visualization methods do not generate any rule-set. Zeiler et al. \cite{zeiler2014visualizing} use similar ideas to analyze what specific kernels in the CNN are invoked. 
There are fewer existing publications on methods for modeling relations between the various important features and generating explanations from them. 
Ferreira et al. \cite{KR2022-45} use multiple mapping networks that are trained to map the activation values of the main network's output to the human-defined concepts represented in an induced logic-based theory. Their method needs multiple neural networks besides the main network that the user has to provide. 

Qi et al. \cite{Qi2021-QIEDN} propose an Explanation Neural Network (XNN) which learns an embedding in  high-dimension space and maps it to a low-dimension explanation space to explain the predictions of the network. A sentence-like explanation including the features is then generated manually. No rules are generated and manual effort is needed. Chen et al.
\cite{chen2019looks} introduce a prototype layer in the network that learns to classify images in terms of various parts of the image. They assume that there is a one to one mapping between the concepts and the kernels. We do not make such an assumption. Zhang et al. \cite{zhang2017growing,zhang2018interpreting} learn disentangled concepts from the CNN and represent them in a hierarchical graph so that there is no assumption of a one to one filter-concept mapping. However, no logical explanation is generated. Bologna et al. \cite{bologna2020two} extract propositional rules from CNNs. Their system operates at the neuron level, while both ERIC and NeSyFold work with groups of neurons. 

Some other works \cite{nesyilp}, \cite{diffilp}, \cite{diffilpext} use a  neurosymbolic system to induce logic rules from data. These systems belongs to the Neuro:Symbolic $\rightarrow$ Neuro category whereas ours belongs to the Neuro;Symbolic category.
%

\section{Conclusion and Future Work}
In this paper we have shown that our NeSyFOLD framework brings interpretability to the image classification task using CNNs. The FOLD-SE-M algorithm cuts-down on the size of rule-set generated significantly. We further show that the semantic labelling algorithm we propose which uses manually annotated semantic segmentation masks of a few images from the \textit{ADE20k} dataset,  leads to highly interpretable rule-sets. We acknowledge that the semantic segmentation masks of images may not be readily available depending on the domain, in which case the semantic labelling of the predicates has to be done manually. Our NeSyFOLD framework helps in this regard as well, as it decreases the number of predicates that need to be labelled. Another option is to find the \textit{top-m} images from the train set according to feature-map norms for each kernel and then obtain semantic segmentation masks for only these images. Notice in our experiment we only used \textit{top-10} from the available images. Moreover, we show that using the sparse kernel learning techniques such as EBP, the rule-set size and consequently number of predicates to be labelled can be further reduced. This also increases interpretability and also saves a lot of manual labelling effort if needed. We plan to evaluate our framework with more sparse kernel learning techniques in the future.

As the number of classes increases, the loss in accuracy also increases due to the binarization of more kernels. Nevertheless, NeSyFOLD outperforms the current SOTA, ERIC in terms of accuracy, fidelity, and rule-set size. We plan to explore end-to-end training of the CNN with the rules generated so that this loss in binarization can be reduced during training itself.

The rules generated by NeSyFOLD are arranged based on the coverage of training images, forming a decision list. Consequently, the topmost rules capture crucial class distinctions. Biases may be apparent in these rules, making them ideal for bias detection. Since the fidelity of the NeSy model is high (at least for less number of classes), experts can review and suggest data changes for training and it is highly likely the retrained model will have lesser bias. It is noteworthy that interpreting a prediction made by the NeSy model becomes easier because of the justification that can be obtained from s(CASP). Hence, having even a large number of rules is less of a problem for an expert scrutinizing a particular prediction of the NeSy model. 

In future, we plan to use NeSyFOLD for real-world tasks such as interpretable breast cancer prediction. Another interesting application is to classify images of unseen classes by constructing rules having predicates from an existing rule-set. This would alleviate the problem of retraining a CNN for classes with slight variations from the current learnt features e.g. classifying beach images by simply writing a rule {\tt target(X, `beach') :- sand(X), water(X).} where {\tt sand} and {\tt water} are pre-learnt predicates.

\bibliography{main.bib}


\appendix
\section*{Acknowledgments}
Authors are supported by US NSF Grants IIS 1910131 and IIP 1916206, US DoD, and various industry grants.

\newpage
\appendix
\centerline{\huge Appendix I}

\bigskip
\noindent In this Appendix we show the performance comparison on various other combinations of classes of the PLACES dataset for NeSyFOLD and NeSyFOLD-EBP. We have also included the labeled rule-sets that NeSyFOLD generates for $1$ run of all the datasets mentioned in the main text. Recall that the generated rule-set along with the trained CNN serves as the NeSy model.

\section{Performace Comparison}
Table \ref{tb_2} shows the performance metrics of NeSyFOLD and ERIC on a subset of the Places dataset.
The $5$ classes chosen are \{desert road(de), driveway(dr), forest road(f), highway(h), street(s)\}. We show the performance by training on a combination of $3$ classes at a time. The other $2$ possible combinations namely, ``dedrh" and ``defs" are shown in the main paper. We chose these particular class combinations because ERIC's performance has been reported on them.  
The hyperparameters used for the results obtained in table \ref{tb_2} as well as the ones reported in the main paper are shown in table \ref{tb_3}. Note these are the \textit{ratio} and \textit{tail} are hyperparameters for the FOLD-SE-M algorithm.

\begin{table*}[t]
\centering
\begin{tabular}{@{}rllllll@{}}

\cmidrule(r){1-7}
\multicolumn{1}{l}{Dataset} & Algo & Fid. & Acc. & \#preds & \#rules & size \\ \cmidrule(r){1-7}
\multirow{3}{*}{dedrf}      & ERIC &- &$0.82$ & $50$ &$31$ &$171$\\
                            & NF   & $0.89 \pm 0.0$ &$\textbf{0.86} \pm \textbf{0.0}$ & $25 \pm 4$ & $20 \pm 2$ & $45 \pm 4$ \\
                            & NF-E &$\textbf{0.90} \pm \textbf{0.02} $ &$ \textbf{0.86} \pm \textbf{0.01} $ & $ \textbf{9} \pm \textbf{4} $ & $ \textbf{10} \pm \textbf{3} $ & $ \textbf{18} \pm \textbf{7} $\\ \cmidrule(lr){1-7}
\multirow{3}{*}{defh}      & ERIC &- &$\textbf{0.81}$ & $48$ &$36$ &$196$\\
                            & NF   & $\textbf{0.85} \pm \textbf{0.01}$ &$0.78 \pm 0.01$ & $34\pm 6$ & $36 \pm 45$ & $82 \pm 13$\\
                            & NF-E &$ \textbf{0.85} \pm \textbf{0.03} $ &$0.78 \pm 0.03$ & $ \textbf{8} \pm \textbf{1} $ & $ \textbf{12} \pm \textbf{1} $ & $ \textbf{22} \pm \textbf{3} $ \\ \cmidrule(lr){1-7}
\multirow{3}{*}{dedrs}      & ERIC & - & $0.76$ & $44$ & $34$ & $183$\\
                            & NF   & $0.90 \pm 0.0$ &$0.88 \pm 0.0$ & $30 \pm 6$ & $21 \pm 3$ & $49 \pm 11$\\
                            & NF-E & $ \textbf{0.91} \pm \textbf{0.01} $ &$ \textbf{0.90} \pm \textbf{0.01} $ & $ \textbf{7} \pm \textbf{2} $ & $ \textbf{9} \pm \textbf{2} $ & $ \textbf{16} \pm \textbf{6} $\\ \cmidrule(lr){1-7}
\multirow{3}{*}{dehs}      & ERIC &- & $0.79$ & $42$ & $36$ & $194$\\
                            & NF   & $0.86 \pm 0.01$ &$0.83 \pm 0.0$ & $28 \pm 5$ & $20 \pm 3$ & $48 \pm 9$\\
                            & NF-E & $ \textbf{0.88} \pm \textbf{0.02} $ &$ \textbf{0.84} \pm \textbf{0.02} $ & $ \textbf{8} \pm \textbf{2} $ & $ \textbf{9} \pm \textbf{2} $ & $ \textbf{15} \pm \textbf{3} $\\ \cmidrule(lr){1-7}
\multirow{3}{*}{drfh}      & ERIC &- & $\textbf{0.81}$ & $47$ & $31$ & $167$\\
                            & NF   & $ \textbf{0.81} \pm \textbf{0.02} $ &$0.76 \pm 0.01$ & $33 \pm 7$ & $33 \pm 7$ & $77 \pm 15$ \\
                            & NF-E & $ \textbf{0.81} \pm \textbf{0.02} $ &$0.76 \pm 0.01$ & $ \textbf{12} \pm \textbf{2} $ & $ \textbf{11} \pm \textbf{2} $ & $ \textbf{24} \pm \textbf{8} $\\ \cmidrule(lr){1-7}
\multirow{3}{*}{drhs}      & ERIC &- & $0.78$ & $47$ & $36$ & $197$\\
                            & NF   & $\textbf{0.83} \pm \textbf{0.01}$ &$\textbf{0.81} \pm \textbf{0.02}$ & $30 \pm 4$ & $21 \pm 5$ & $52 \pm 11$ \\
                            & NF-E & $ \textbf{0.83} \pm \textbf{0.03} $ &$0.80 \pm 0.03$ & $ \textbf{10} \pm \textbf{5} $ & $ \textbf{10} \pm \textbf{4} $ & $ \textbf{19} \pm \textbf{11} $\\ \cmidrule(lr){1-7}
\multirow{3}{*}{fhs}      & ERIC &- & $0.72$ & $56$ & $34$ & $185$ \\
                            & NF   & $0.87 \pm 0.0$ &$0.82 \pm 0.0$ & $21 \pm 4$ & $17 \pm 1$ & $38 \pm 4$\\
                            & NF-E &$ \textbf{0.90} \pm \textbf{0.0} $ &$ \textbf{0.84} \pm \textbf{0.0} $ & $ \textbf{11} \pm \textbf{2} $ & $ \textbf{11} \pm \textbf{1} $ & $ \textbf{22} \pm \textbf{2} $\\ \cmidrule(lr){1-7}
\multirow{3}{*}{drfs}      & ERIC &- & $0.70$ & $47$ & $33$ & $181$\\
                            & NF   & $0.87 \pm 0.01$ &$0.85 \pm 0.01$ & $26 \pm 6$ & $20 \pm 3$ & $46 \pm 11$\\ 
                            & NF-E &$ \textbf{0.89} \pm \textbf{0.01} $ &$ \textbf{0.86} \pm \textbf{0.01} $ & $ \textbf{13} \pm \textbf{1} $ & $ \textbf{14} \pm \textbf{1} $ & $ \textbf{27} \pm \textbf{4} $\\ \cmidrule(lr){1-7}
    
\end{tabular}
\caption{Comparison of ERIC vs NeSyFOLD (NF) vs NeSyFOLD-EBP (NF-E)}
\label{tb_2}
\end{table*}

\begin{table}
\centering
\begin{tabular}{lrrrr}
    \toprule
    Dataset &Ratio & Tail \\
    \cmidrule(lr){1-3}
    GT43 &5 & 1e-3  \\ 
    P2 &0.8 & 5e-3\\ 
    P3.1 &0.8 & 5e-3\\ 
    P3.2 &0.8 & 5e-3\\
    P3.3 &0.8 & 5e-3\\
    P5 &5 & 4e-3\\
    P10 &10 & 5e-3\\ 
    dedrf &0.8 & 5e-3\\ 
    dedrs &0.8 & 5e-3\\ 
    defh &5 & 5e-3\\ 
    dehs &0.8 & 5e-3\\ 
    drfh &5 & 5e-3\\ 
    drfs &0.8 & 5e-3\\ 
    drhs &0.8 & 5e-3\\ 
    fhs &0.8 & 5e-3\\
    \bottomrule
\end{tabular}
\caption{Hyperparameter values for NeSyFOLD and NeSyFOLD-EBP}
\label{tb_3}%
\end{table}


\section{Labelled rule-sets}
In this section, we show all labelled rule-sets for the datasets used in the main paper for the experiments. All these rule-sets were generated using NeSyFOLD and NeSyFOLD-EBP. These rule-sets along with the trained CNN serve as the NeSyFOLD model and are used for prediction. All these are class combinations of the Places dataset. Note that as the classes increase the accuracy of both NeSyFOLD and NeSyFOLD-EBP drops hence the rule-sets also start to make less intuitive sense.

\medskip\noindent\textbf{PLACES2 \{``bathroom", ``bedroom"\}:}\\
The hyperparameter values used are $ratio = 0.3$, $tail = 5e^{-3}, margin = 0.05$\\
\small{
\begin{verbatim}
VANILLA:
target(X,`bedroom') :- not mirror1(X), bed4(X).
target(X,`bathroom') :- countertop1(X).
target(X,`bedroom') :- bed7(X), not ab1(X).
target(X,`bathroom') :- not bed3(X).
target(X,`bedroom') :- not bed3(X).
target(X,`bedroom') :- not bathtub1(X), bed6(X).
target(X,`bedroom') :- bed2(X).
target(X,`bathroom') :- mirror1(X).
ab1(X) :- not bed1(X), bathtub1(X), not bed5(X).
\end{verbatim}
}

\small{
\begin{verbatim}
EBP:
target(X,'bedroom') :- not sink1_wall1_toilet1(X), 
    not ab1(X), not ab2(X).
target(X,'bathroom') :- not bed1(X).
target(X,'bathroom') :- not bed2_floor1(X).
ab1(X) :- not bed1(X), wall4(X), not bed3(X).
ab2(X) :- wall3(X), not wall2(X).
\end{verbatim}

}
\medskip\noindent\textbf{P3.1 \{``bathroom", ``bedroom", ``kitchen"\}:}\\
The hyperparameter values used are,
$ratio = 0.8$, $tail = 5e^{-3}, margin = 0.05$
\small{
\begin{verbatim}
VANILLA:
target(X,`kitchen') :- cabinet1_wall1(X),
    not ab4(X), not ab5(X).
target(X,`bedroom') :- bed6(X), not ab6(X).
target(X,`bathroom') :- sink1_toilet1_wall5(X),
    not ab7(X).
target(X,`bathroom') :- not cabinet3(X),
    wall15(X), not ab9(X).
target(X,`bedroom') :- bed12(X), not ab10(X),
    not ab12(X).
target(X,`kitchen') :- cabinet2(X), not ab13(X).
target(X,`bathroom') :- sink2_wall7(X),
    not bed10(X).
target(X,`kitchen') :- wall14(X), not ab14(X).
target(X,`bathroom') :- not bed5(X), not wall9(X),
    wall11_sink3(X).
target(X,`bedroom') :- bed1(X).
target(X,`kitchen') :- cabinet5(X), not wall8(X).
target(X,`bathroom') :- not bed9(X), cabinet5(X).
ab1(X) :- not bed7(X), cabinet2(X).
ab2(X) :- bed3(X), not ab1(X).
ab3(X) :- cabinet9_wall13(X), not ab2(X).
ab4(X) :- not cabinet4(X), not ab3(X).
ab5(X) :- not cabinet7(X), not cabinet6_wall12(X),
    sink1_toilet1_wall5(X).
ab6(X) :- wall11_sink3(X), sink1_toilet1_wall5(X).
ab7(X) :- not sink4(X), cabinet9_wall13(X),
    not wall15(X).
ab8(X) :- not sink1_toilet1_wall5(X), not bed8(X),
    not wall4_bed4(X).
ab9(X) :- cabinet9_wall13(X), not ab8(X).
ab10(X) :- not bed11(X), wall3(X).
ab11(X) :- not cabinet8(X), not sink4(X).
ab12(X) :- not bed2(X), not bed3(X), not ab11(X).
ab13(X) :- not cabinet4(X), not cabinet8(X),
    not wall2(X).
ab14(X) :- not wall10_toilet2_floor1(X), 
    not wall6(X).
\end{verbatim}
}

\small{
\begin{verbatim}
EBP:
target(X,`kitchen') :- cabinet2(X), not ab1(X).
target(X,`bedroom') :- bed1(X).
target(X,`bathroom') :- sink1_toilet1(X).
target(X,`bathroom') :- wall2(X).
target(X,`kitchen') :- cabinet1(X).
target(X,`bedroom') :- bed2(X).
target(X,`bathroom') :- wall1(X).
target(X,`kitchen') :- 
    work_surface1_kitchen_island1(X).
ab1(X) :- not cabinet1(X), bed1(X).
\end{verbatim}
}

\medskip\noindent\textbf{P3.2 \{``desert road", ``forest road", ``street"\}:}
The hyperparameters values used are, 
$ratio = 0.8$, $tail = 5e^{-3}, margin = 0.05$
\small{
\begin{verbatim}
VANILLA:
target(X,`street') :- building11(X).
target(X,`forest_road') :- building9(X),
    not ab1(X), not ab2(X).
target(X,`desert_road') :- building5(X),
    not ab3(X).
target(X,`forest_road') :- not building4(X),
    not building9(X), not ab4(X), not ab5(X).
target(X,`desert_road') :- not building12(X),
    not building6(X), not building5(X),
    not building8(X).
target(X,`street') :- building13(X).
target(X,`forest_road') :- not building1(X),
    not building8(X).
target(X,`desert_road') :- building7(X).
target(X,`street') :- not building13(X),
    building3(X).
target(X,`forest_road') :- building8(X).
ab1(X) :- building2(X), building15(X).
ab2(X) :- not building10(X),  road1(X), road2(X).
ab3(X) :- not building14(X), building6(X).
ab4(X) :- building10(X), not building15(X).
ab5(X) :- tree1(X), not road1(X).

\end{verbatim}

\small{
\begin{verbatim}
EBP:
target(X,'street') :- building1(X).
target(X,'forest_road') :- trees1(X).
target(X,'desert_road') :- sky2(X).
target(X,'forest_road') :- tree1(X).
target(X,'desert_road') :- not building2(X),
    sky1(X).
target(X,'street') :- building2(X).
\end{verbatim}
}
\medskip\noindent\textbf{P3.3 \{``desert road", ``driveway", ``highway"\}:}
The hyperparameters values used are, 
$ratio = 0.8$, $tail = 5e^{-3}, margin = 0.1$
\small{
\begin{verbatim}
VANILLA:
target(X,`highway') :- not road2_ground1(X),
    not ab4(X), not ab5(X).
target(X,`driveway') :- house4(X), not ab7(X).
target(X,`desert_road') :- not house3(X),
    not ab10(X), not ab11(X),  not ab12(X).
target(X,`highway') :- road12(X).
target(X,`driveway') :- not road20(X),
    not road18(X), not house4(X).
target(X,`highway') :- not road12(X), not ab13(X).
target(X,`desert_road') :- not building1(X),
    not grass1(X).
target(X,`driveway') :- not ground2(X).
ab1(X) :- not road11(X), not road1(X).
ab2(X) :- not road21(X), road11(X), not road17(X),
    trees1(X).
ab3(X) :- not house4(X), not ab1(X), not ab2(X).
ab4(X) :- not road19(X), not ab3(X).
ab5(X) :- ground2(X), building1(X).
ab6(X) :- not road10(X), not road19(X), house3(X).
ab7(X) :- not house1(X), road7(X), not ab6(X).
ab8(X) :- not road6(X), road16(X), not road8(X).
ab9(X) :- road13(X), not road4(X), not house2(X).
ab10(X) :- not mountain1(X), not road3(X), not ab8(X),
    not ab9(X).
ab11(X) :- road14(X), not road15(X).
ab12(X) :- not road9(X), trees2(X), not road16(X).
ab13(X) :- ground2(X), not road5(X).

\end{verbatim}
}
\small{
\begin{verbatim}
EBP:
target(X,'highway') :- road2(X), not ab1(X),
    not ab2(X).
target(X,'driveway') :- house2(X).
target(X,'desert_road') :- sky2(X), not ab3(X).
target(X,'highway') :- road4(X), not ab4(X).
ab1(X) :- sky2(X), not grass1(X).
ab2(X) :- house2(X), house1(X).
ab3(X) :- not sky1(X), road3(X).
ab4(X) :- not road1(X), not sky2(X), road2(X).
\end{verbatim}
}
\medskip\noindent\textbf{PLACES5 \{``bathroom", ``bedroom", ``kitchen", ``dining room", ``living room"\}:}\\
The hyperparameter values used are,
$ratio = 5$, $tail = 4e^{-3}, margin = 0.05$
\small{
\begin{verbatim}
VANILLA:
target(X,`living_room') :- window4(X), not ab4(X),
    not ab12(X), not ab13(X).
target(X,`dining_room') :- cabinet4(X), not ab16(X),
    not ab18(X).
target(X,`kitchen') :- chair4(X), not ab20(X),
    not ab21(X).
target(X,`bedroom') :- bed7(X), not ab23(X),
    not ab27(X).
target(X,`bathroom') :- countertop1(X), not ab29(X).
target(X,`living_room') :- not mirror3(X),
    not ab36(X), not ab38(X), not ab40(X).
target(X,`kitchen') :- table3(X), not ab45(X),
    not ab48(X).
target(X,`bathroom') :- cabinet6(X), not window1(X),
    bathtub2(X).
target(X,`bedroom') :- bed18(X), not window1(X),
    not ab50(X).
ab1(X) :- not chair3(X), armchair2_window5(X).
ab2(X) :- not bed9(X), not ab1(X).
ab3(X) :- sofa3(X), not bed9(X), not ab2(X).
ab4(X) :- not window3(X), not ab3(X).
ab5(X) :- not sink1_table1(X), not chair8(X).
ab6(X) :- not bed5(X), cabinet8(X), not ab5(X).
ab7(X) :- bed2(X), not chair9(X).
ab8(X) :- chair1(X), not sofa6(X).
ab9(X) :- sofa2(X), not ab6(X), not ab7(X),
    not ab8(X).
ab10(X) :- sofa5(X), bed2(X).
ab11(X) :- not sofa2(X), sofa5(X), not ab10(X).
ab12(X) :- window3(X), not ab9(X), not ab11(X).
ab13(X) :- sink1_table1(X), bed17(X).
ab14(X) :- cabinet7(X), mirror2_shower_screen1(X).
ab15(X) :- not bathtub1(X), not bed9(X),
    cabinet7(X), not ab14(X).
ab16(X) :- not cabinet9(X), not ab15(X).
ab17(X) :- not countertop2(X), not bed10(X).
ab18(X) :- not cabinet5(X), not ab17(X).
ab19(X) :- table3(X), sofa7_cabinet10_chair10(X).
ab20(X) :- not chair8(X), not ab19(X).
ab21(X) :- bed5(X), bed4(X).
ab22(X) :- bed9(X), bed3(X).
ab23(X) :- cabinet11(X), not ab22(X).
ab24(X) :- not bathtub1(X), bed10(X).
ab25(X) :- armchair1(X), not ab24(X).
ab26(X) :- not cabinet11(X), not ab25(X).
ab27(X) :- not bed9(X), not ab26(X).
ab28(X) :- not chair9(X), not bed13(X).
ab29(X) :- not mirror1_sink2(X), not ab28(X).
ab30(X) :- not sofa2(X), chair2(X).
ab31(X) :- not bed12(X), not sofa2(X),
    not ab30(X).
ab32(X) :- chair6(X), not bed12(X), not ab31(X).
ab33(X) :- not bed1(X), bed11(X).
ab34(X) :- cabinet3(X), not ab33(X).
ab35(X) :- armchair2_window5(X), cabinet3(X),
    not ab34(X).
ab36(X) :- not sofa3(X), not ab32(X), not ab35(X).
ab37(X) :- not sofa1(X), not bed15(X).
ab38(X) :- sofa3(X), not sofa1(X), not ab37(X).
ab39(X) :- chair6(X), not table2(X).
ab40(X) :- not picture1(X), chair6(X), not ab39(X).
ab41(X) :- not chair1(X), not sofa4(X).
ab42(X) :- sofa4(X), bed8(X).
ab43(X) :- not bed8(X), cabinet1(X).
ab44(X) :- not cabinet2(X), not ab41(X),
    not ab42(X), not ab43(X).
ab45(X) :- not window2(X), not ab44(X).
ab46(X) :- chair5(X), chair7(X).
ab47(X) :- bed6(X), chair5(X), not ab46(X).
ab48(X) :- window2(X), bed6(X), not ab47(X).
ab49(X) :- bed16(X), bed1(X).
ab50(X) :- not window1(X), not bed14(X),
    not ab49(X).
\end{verbatim}
}
\small{
\begin{verbatim}
EBP:
target(X,`living_room') :- sofa1(X),
    not sink1_wall1_toilet1(X), not ab2(X),
    not ab3(X).
target(X,`dining_room') :- cabinet1(X), not ab8(X).
target(X,`kitchen') :- chair2(X), not ab10(X).
target(X,`bedroom') :- bed2(X).
target(X,`bathroom') :- sink1_wall1_toilet1(X).
target(X,`living_room') :- 
    armchair1_floor1_bed4(X), not table1(X).
target(X,`kitchen') :- table3_chair4(X),
    not chair2(X), not ab11(X).
ab1(X) :- not bed3(X), chair3(X).
ab2(X) :- bed2(X), not ab1(X).
ab3(X) :- chair2(X), not bed1(X).
ab4(X) :- not wall2(X), not cabinet2(X).
ab5(X) :- not chair3(X), not ab4(X).
ab6(X) :- not cabinet3(X), not ab5(X).
ab7(X) :- not sink1_wall1_toilet1(X),
    not bed2(X), not chair3(X), not ab6(X).
ab8(X) :- not wall3_cabinet4(X), not ab7(X).
ab9(X) :- not bed2(X), chair1(X).
ab10(X) :- sofa1(X), not ab9(X).
ab11(X) :- not table2(X), not chair1(X).
\end{verbatim}
}

\end{document}